# Fractal Language Modelling
## by Universal Sequence Maps (USM)

Jonas S Almeida[1*], Daniel E Russ[1], Susana Vinga[2,3], Ines Duarte[1,2], Lee Mason[1], Prahulla Bhawsar[1,5], Aaron Ge[6], Arlindo Oliveira[4], Jeya B Balasubramanian[1].

* jonas.dealmeida@nih.gov
1) National Cancer Institute, Division of Cancer Epidemiology and Genetics (DCEG), NCI Shady Grove, Rockville, MD 20850, USA.
2) INESC-ID, Instituto Superior Técnico, Universidade de Lisboa, 1000-029 Lisbon, Portugal.
3) IDMEC, Instituto Superior Técnico, Universidade de Lisboa, 1049-001 Lisbon, Portugal.
4) Instituto de Engenharia de Sistemas e Computadores, Instituto Superior Técnico, University of Lisbon, Lisbon 1000-029, Portugal.
5) Department of Biomedical Informatics, Stony Brook Medicine, 100 Nicolls Rd, Stony Brook, NY 11794, USA.
6) Institute of Health Computing at the University of Maryland, School of Medicine, Baltimore MD 21201,USA.

*This project was funded by the Division of Cancer Epidemiology and Genetics (DCEG), National Cancer Institute, NIH, USA, reference # Episphere 10901*

## Abstract

**Motivation** - With the advent of Language Models using Transformers, popularized by ChatGPT, there is a renewed interest in exploring encoding procedures that numerically represent symbolic sequences at multiple scales and embedding dimensions. The challenge that encoding addresses is the need for mechanisms that uniquely retain contextual information about the succession of individual symbols, which can then be modeled by nonlinear formulations such as neural networks.
**Context** - Universal Sequence Maps(USM) are iterated functions that bijectively encode symbolic sequences onto embedded numerical spaces. USM is composed of two Chaos Game Representations (CGR), iterated forwardly and backwardly, that can be projected into the frequency domain (FCGR). The corresponding USM coordinates can be used to compute a Chebyshev distance metric as well as *k-mer* frequencies, without having to recompute the embedded numeric coordinates, and, paradoxically, allowing for non-integers values of *k*.
**Results** - This report advances the bijective fractal encoding by Universal Sequence Maps (USM) by resolving seeding biases affecting the iterated process. The resolution had two results, the first expected, the second an intriguing outcome: 1) full reconciliation of numeric positioning with sequence identity; and 2) uncovering the nature of USM as an efficient numeric process converging towards a steady state sequence embedding solution. We illustrate these results for genomic sequences because of the convenience of a planar representation defined by an alphabet with only 4 tokens (the 4 nucleotides). Nevertheless, the application to alphabet of arbitrary cardinality was found to be straightforward.

**Availability**: Supplementary interactive material can be found at the following locations:
a) Fold Web Application with improved USM mapping: https://usm.github.io/fold
b) Interactive tabulation of USM recurrence (Fig 6): https://usm.github.io.
c) Demonstration of generalization to non-genomic alphabets (Fig 8): https://observablehq.com/@episphere/usm.

# Introduction

The popularity of Chaos Theory in the 90's led to the exploration of graphical representations of symbolic sequences using iterated functions, such as Chaos Game Representation (CGR) of genomic sequences [1]. The iterated mapping between symbolic and numeric domains was initially interpreted as a redundant representation of a Markov chain, using fixed-length L-tuple frequencies to calculate transition probabilities [2]. However, in the early 2000's it was found that, quite the contrary, CGR was the more generic representation of sequence succession [3], leading to the development of alignment-free sequence analysis methods [4]. Subsequent work led to additional discoveries[5], including the critical bijective mapping property, the implicit alignment of bidirectional encoding[6], the ability to decompose sequence comparison with efficient map-reduce operations and an exact distance metric [7], as recalled here, in the Methods section.

A decade later, the development of convolutional neural network methodologies for image analysis found application to biological sequence analysis using the graphical representation of the iterated map as the pictorial form of L-tuple sequence decomposition[8]. That representation of token succession as a frequency table is sometimes described as a "frequency Chaos Game Representation", or FCGR for short[9] provides a convenient input format for encoding generative language modeling that benefits from the scale-independence of alignment-free sequence analysis methodologies[4]. FCGR, and by extension the bijective USM mapping that produces it, act like a fractal k-mer transition table, where the length, *n*, does not have to be an integer[6][7]. This combination of a bijective map between embedded numeric coordinates and sequences of symbols, make the USM iterated map a compelling numeric representation where to investigate the properties of the corresponding symbolic space.

The advent of ChatGPT in the early 2020's shed new light over the critical role of positional encoders in the generative modeling of token emission, this time using Transformers as encoders instead of recurrent neural networks [10]. Several computational genomics challenges benefit from these advances, such as genotype imputation[11][12] and variant effects[13]. The positional encoding followed by the self-attention mechanism used by Transformers, in particular, was found to be particularly suitable to the modeling of genomes in several studies[14], including genotype imputation[15].

Generative models approach embedding input data by indexing the context of each emission, to enable sampling of both proximal (for example the preceding token) and distal (the value of the token a few positions away) context patterns[16]. The scale-free (k-mer length independent) representation of sequence succession) afforded by iterated USM maps[5] may offer a different approach to computational genomics through the development of language models [17][18].

The starting point for results reported below is to recall that a language model[19] is a function that assigns transition probabilities to each possible token, $T_{i=n+1}$, given a sequence of *n* token observations $T_{i=1...n}$. That generative function can be as simple as a L-tuple transition table, or as complex as a deep learning neural network with a transformer architecture in the case of Large Language Models (LLM). Either way, the scalability of the generative model will need an efficient unsupervised embedding mechanism describing "a generalized conservation score" [13], as illustrated by the Genomic Pretrained Network approach (GPN) to learning genome-wide genomic variability. In this report, we take the minimalist route to the representation of conserved variability by exploring variations on the iterated Universal Sequence Map technique[6] as the encoder, making the most of the scale independence and bijective mapping properties of that generative embedded space [5].

## Methods

The proposed multi-scale (scale-free) solution to sequence embedding is based on the iterated map generated by Chaos Game Representation (CGR). As outlined in the Introduction, the original CGR specific goal was the visualization of genomic sequences (Figure 1) in a planar unit square[20]. In the original procedure, each corner of a unit square is assigned to a nucleotide, and, starting with the middle position [½ ,½], each coordinate is moved by half the distance towards the corner assigned to the corresponding nucleotide. The procedure is detailed below, but is best understood by using the accompanying Fold Web tool (see under Availability). We'll now expand the CGR notation to create the new bidirectional, scale-free, USM embedding object. The construction will be illustrated for the DNA sequence "*GATTACA*". For each symbolic sequence, *S*, with *n* units, where $S=s_1...s_n$, such that $s_i \in A$; with *Abc* being the alphabet of possible *m* units, each assigned to a corner of the unit hypercube with dimension $h = ceil(log_2(m))$. For example, for the DNA sequence *S='GATTACA', n= 7, Abc=['A','C','G','T'], m=4, and h=2* (the unit hypercube is a planar unit square) with corner coordinates "A": [0, 0]; "C": [0, 1]; "G": [1, 0]; "T": [1, 1], For convenience, the coordinates of each of the *m* corners, $E=E_{i=1...m, j=1...h}$ will be denoted by the corresponding nucleotide. Recall "Edges" and "corners" will be used interchangeably in this report. For example, $E_{4,2}$ indicates the y-axis coordinate of the 4th corner corner, "T", and the second which has a value of *1*. Essentially, $E$ serves as a lookup table that maps each symbol $s_i$ to a corner of the unit hypercube and returns the coordinates of that corner, $E(s_i) \in \mathbb{R}^h$. With this notation in mind, the forward encoding for the original CGR procedure, starting with a seed of [½ ,½] , will generate a two-dimensional map (h=2), $U=u_{1,1}...u_{h,n}$ , where each of the two numeric coordinate arrays is 7 units long (*n=7*), and where $u_{j=1,...,h, i=1,...,n} \in [0, 1]$, are iterated as described in equation 1, leading to the results in Table 1. The interactive USM graphic (Figure 1) can be generated using the accompanying interactive *Fold* web application.(Figure 1).

$$u_{j,i} = u_{j,i-1} + (E_{j,i} - u_{j,i-1})/2 \text{ , with } j = 1...h \text{ and } i = 1...n \qquad \text{(Equation 1)}$$

**Table 1** - Numeric detail on the execution of equation 1 by running `u = new usm('GATTACA')`. The numerical coordinates can be explored by inspecting `u.forward` and `u.backward`, which are depicted graphically in the interactive Figure 1. The mapping procedure described in Equation 1 starts by finding the unique letters of the alphabet, `u.abc`, which are then automatically assigned to corners to the USM object used by both CGRs.

| | |
|---|---|
| As per Equation 1, "GATTACA" is mapped forwardly to<br>$u_1$ = [0.75, 0.375, 0.6875, 0.84375, 0.421875, 0.2109375, 0.10546875]<br>$u_2$ = [0.25, 0.125, 0.5625, 0.78125, 0.390625, 0.6953125, 0.34765625]<br>and backwardly to<br>$u_1$ = [0.69140625, 0.3828125, 0.765625, 0.53125, 0.0625, 0.125, 0.25]<br>$u_2$ = [0.20703125, 0.4140625, 0.828125, 0.65625, 0.3125, 0.625, 0.25] | u.edges<br>{ A: [0,0],<br>C: [0,1],<br>G: [1,0],<br>T: [1,1]  } |

The live results can be reproduced in a regular web browser with no need for downloads or installations, by using the accompanying open source JavaScript library. This library supports both the assembly of USM objects for given sequences, and the interactive graphics depicted in Figure 1. They can also be explored directly in the Web Browser's native console, which Web Browsers provide under "Developer Tools". The console can be used straightforwardly as any console application by loading the open source library:

1. Load USM ES6 module:
   *usm = (await import(`https://usm.github.io/3/usm.mjs`)).USM*
2. Create an instance of a Universal Sequence Map (USM) for the sequence `GATTACA`
   *u = new usm(`GATTACA`)*
3. Explore the *u* instance of "GATTACA" by investigating its attributes and methods, starting with USM's forward and backward embeddings, *u.forward* and *u.backward*, followed by the coordinates for the corners, at *u.edges*.

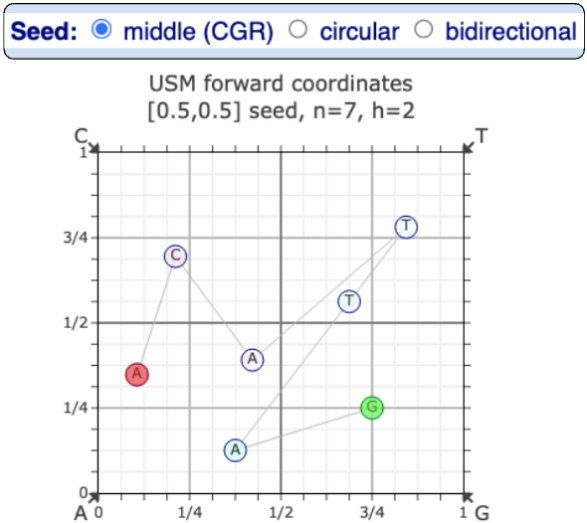

**Figure 1** - snapshot of using the accompanying Fold Web App (usm.github.io/fold ) to explore the CGR procedure being applied to "GATTACA" for all three seeding solutions (select corresponding radio button): Middle (original CGR), Circular; and Bidirectional. Note how, as described in Equation 1, each nucleotide position transition points to the corner of the corresponding nucleotide: "GA" goes half the distance between the current position of "G" and the corner of "A". The original unidirectional fixed-seed CGR procedure was also expanded to include a backward mapping procedure. Together, the two iterations, forward and backward, define a Universal Sequence Map (USM), supporting the calculation of sequence similarity (Equation 6) without the need for dynamic programming [6,21].

# Results

The USM encoder described in this report advances the original USM approach by resolving corner-edge effects ("corner" and "edge" will be used here interchangeably) that create inconsistencies in the recursive numerical process. Without these improvements, the conventional CGR's static mapping will be degraded by corner effects at the beginning and end of sequence, fundamentally hindering the appeal of USM as a coherent, general purpose, language model. For example, a DNA sequence composed of a single nucleotide will not be projected to the corresponding corner of the embedded space (see accompanying Fold Web Application). Instead, the static CGR seeding will start at the midpoint of that nucleotide's quadrant. This inconsistency is typically ignored by noting that it doesn't affect long sequences significantly. That is indeed the case, but then it affects processing of large collections of small sequences such as those generated by shotgun sequencing. Furthermore, seeding with the original CGR method (static seeding with position ½ for all dimensions), or, it turns out, with any arbitrary static seed, has the potential to leave behind distortions affecting the binning of CGR coordinates determining k-mer frequency (FCGR)[22]. This Results section will first resolve the seeding procedure as a dynamic numeric process, and will then use those results to generate the corresponding scale-free frequency tables.

## Seeding Universal Sequence Maps (USM)

Universal Sequence Maps[6] are efficient[21] iterated mapping techniques projecting symbolic sequences into numeric spaces bijectively (one-to-one). This projection is also bidirectional - it maps the sequence bijectively forwardly, and bijectively backwardly (see Methods). However, we found that there are inconsistencies in the original CGR mid point seeding that need to be addressed in order to generate a coherent sequence map, regardless of the application or the sequence composition. Firstly, at the root of the proposed solution is the realization that the USM mapping needs to be computed as a dynamic process that converges towards a coherent numeric solution. The second realization is that avoiding corner effects requires seeding the USM iterated mapping procedure differently. The resulting three seeding solutions (Figure 2) are best understood by using the accompanying web application (see Fold Web Application in Availability).

Using the midpoint as the seed for the Chaos Game Representation procedure (CGR) has the paradoxical property of starting the encoding with a position that cannot itself be decoded. To reach the USM map midpoint (½) iteratively would require a long sequence segment filled with the same type of nucleotide, longer than the numeric resolution of the embeddings coordinates, followed by a new nucleotide type. This can be visualized by mapping, for example, "AAAAAAAAAAAT" using the Fold Web Application. Making matters worse, that exercise will reflect the propagation of the distortion to the FCGR binning. The existence of tail effects on either end of a sequence makes it glaringly apparent that USM is best conceived as a numeric method, along the lines of the circular seeding procedure for short sequences[7]. As illustrated in Figure 2 of that report[7] and recalled here also in Figure 2, the circular encoding procedure uses the tail of the backward direction to seed the head of the forward direction of the numerical iteration, which will run until the values of the feature vectors converge. Similarly, the new bidirectional dynamic seeding procedure is created simply by circulating by alternating directions (Fig 2). While for circular DNA sequences (plasmids, many viral genomes such as bacteriophages, circRNA, etc) circular iteration seeding is the natural solution, we have subsequently observed less obvious tail effects when encoding longer sequences with different nucleotide compositions between the two ends. To address biases more broadly, while preserving the coherence of decoding corners of quadrants and subquadrants accurately, we introduced this new bidirectional encoding procedure where the iteration of one direction is seeded by the coordinates for the same nucleotide in the alternate direction. All 3 seeding solutions, CGR, circular and bidirectional were implemented in the interactive Fold Web application (see Availability), and are described graphically in Figure 2 and 3.

## Iterated seeding

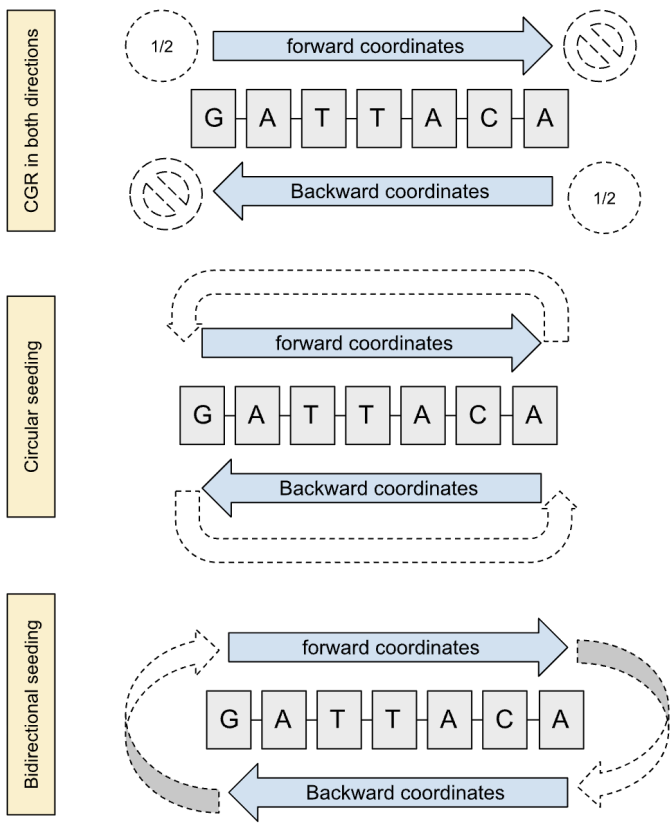

**Figure 2** - Seeding approaches used by USM positional encoders in chronological order, from top to bottom. In 2002 CGR procedure (single pass) was used to generate bidirectional USM coordinates (forward and backward). A sequence similarity metric was then identified that does not require alignment or dynamic programming[6]; in 2012 a variant of the iterated method was proposed as a recursive process that converges for sequences of any size and composition, including a single nucleotide sequences[7]. In this report, we advance the alternate bidirectional USMfold procedure, which switches the seeding direction between encoding rounds (forward and backward), to accommodate shifts in genome composition: when encoding reaches the tail of a sequence, it continues back from the tail in the alternate direction. Interestingly, a close inspection of alternating bidirectional seeding in this figure progresses as if on a Möbius strip, which, oddly, can occur naturally[23].

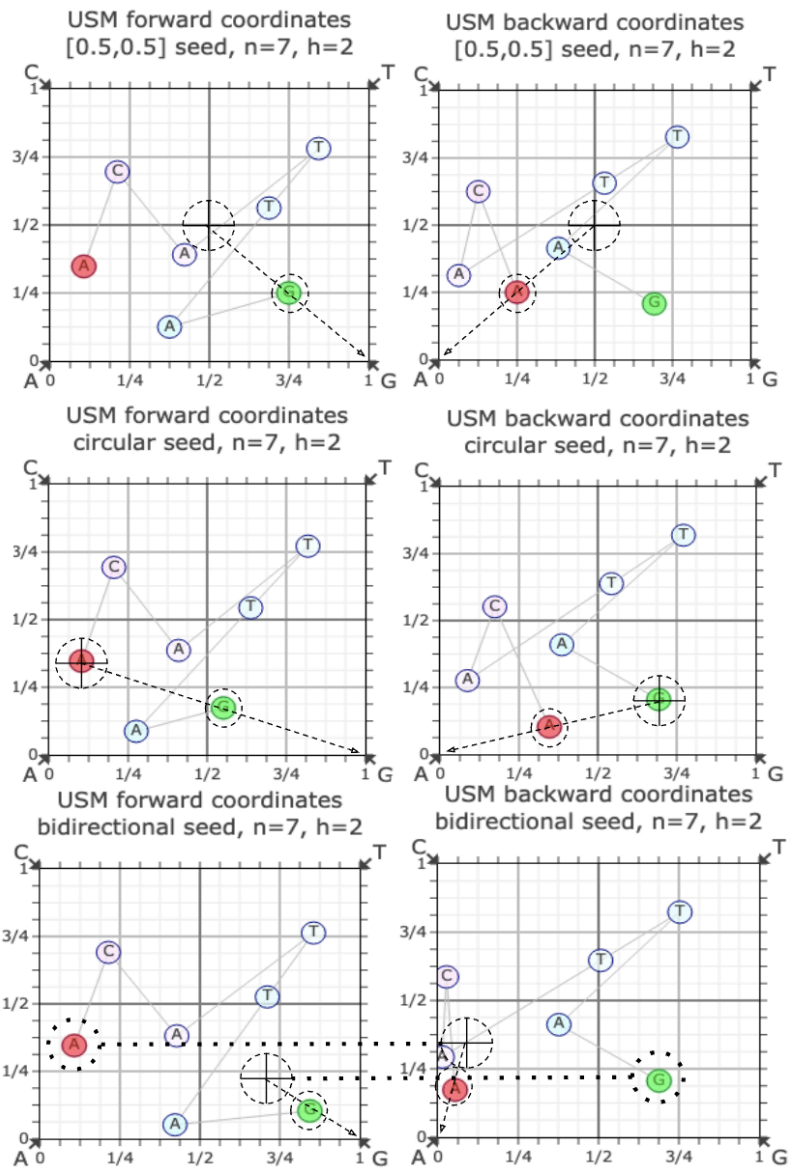

**Figure 3** - Using "GATTACA" as an example, this figure details the seeding process and the distinction between one pass iteration of the fixed CGR seed (½), and the recursive iteration taking place with circular and bidirectional seeding, where the coordinate values converge to the final positions reported for the USM map. The three methods are laid out in the same vertical order as the diagram in Figure 2, with the first column showing the iterated solution for forward mapping, and the second column showing the coordinates for the backward iteration. For each of the 6 plots the basic iteration is to move the nucleotide sequence coordinates half the distance to the corresponding nucleotide's edge. In each case, regardless of the coordinates being iterated forward or backward, the first nucleotide coordinates are highlighted in green, and the last nucleotide is highlighted in red. The seed of each of the 6 maps is marked with a crosswire within a large circle, and the solution for the next coordinate (the first sequence unit) is indicated by a smaller dashed circle, at half the distance from the seed and the corresponding unit edge. As elaborated in the text, all three procedures define USM maps, each with the same alignment-free properties[6] of no need for dynamic programming and a shared distance metric, as described in the original 2012 report [7], reviewed for the wider context of iterated maps two years later[5]. These figures can be reproduced using the accompanying web tool (see Availability).

## Scale-free frequency tables

As explored above, getting the iterative seeding right is important for the sake of coherence, but it should be noted that for long sequences the difference between the 3 methods is numerically insignificant except for the first and last hundred positions (the exact value depends on the sequence composition and numerical resolution of the computational environment).

The bijective and bidirectional mapping of symbolic sequences to a USM unit hypercube (a square for nucleotide sequences, where $h=2$) has important logistic advantages[7]. The improvement can be appreciated by using the Fold Web Application to calculate all k-mer frequencies of 8 nucleotide oligomers ($k=8$), for the 244,589 nucleotide-long sequence of the EGFR gene[24]. That sequence, NCBI id 399923581, is the default input sequence of this Web Application, which is loaded, processed in real time, in-browser, by the Fold Web Application, to generate the FCGR plots in (Figure 4). In a moderately resourced laptop, this iterated indexing of a quarter million sequence takes approximately 300 milliseconds, which is then binned for the count for all possible, 8 nucleotide long sequences, which corresponds to a $2^8 \times 2^8 = 65536$ bin array), in about another 10 milliseconds. This frequency array, sometimes described as a frequency CGR (FCGR)[21][22], is represented as a colorscale density map. The reader is encouraged to use their own genomic sequence data and change the processing options to explore the advantages, and limits of the USM approach.

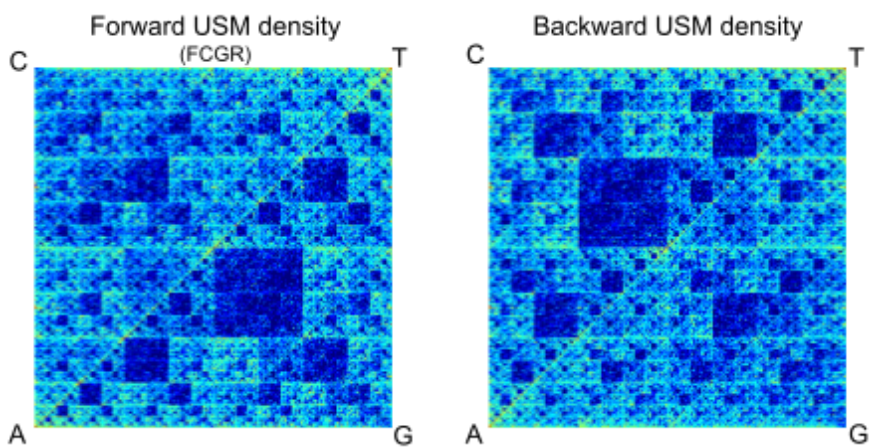

**Figure 4** - Density plot for the 244,589 long sequence of the EGFR gene, resolved for 8 oligomer frequencies (binning all the occurrences of 8 nucleotide long sequences). An interactive tool is provided (see Availability) where other parameter values can be used. By default, bidirectional seeding was used to produce these plots. Note close inspection will reveal that forward and backward densities are not symmetrical.

This density map, which can be constructed for any vocabulary cardinality, offers a straightforward representation for use as input for machine learning, such as neural based architectures[22]. In the work reported here, the focus will remain on the parameterization of the USM encoding (Fig 1,2,3), followed by the frequency count (Figure 4) to generate k-mer FCGR density tables. In the USM procedure, this can also be performed in both forward and backward directions.

## USM Embedded Metric Space

The identification of a metric distance function for the embedded space created by USM coordinates was driven by the opportunity to devise analytic methods that rely on the numeric coordinates in the USM embedded space. The first clue that such a metric should be possible as an exercise if fractal geometry is the folding pattern where, regardless of the position of an individual nucleotide in the USM map, the transition to the next 4 possible nucleotide type define a square of fixed size ½ x ½ (Figure 5). The second clue was that, if two sequence units positioned in the same USM transition towards the same corner (towards the same type of nucleotide), the Euclidean distance between the two is halved for each additional similar unit added to both aligned sequences. However, using Euclidean distance to derive a new metric for the USM map turned out to be a dead-end: many pairs of positions will be at short Euclidean distances from each other simply because they are adjacent to the same boundary between quadrants. The only guarantee when comparing two positions in the USM space is that the number of consecutive transitions to a similar corner (alignment) is no less than $-log_2$ of that Euclidean distance. For example, if two positions in the USM space are at an Euclidean distance of *0.0123* from each other, they are aligned in a segment of at least *6* units. Three realizations were therefore apparent: a) Euclidean distances will not do; b) a metric distance function would need to operate by folding quadrants if it was to make the most of USM's bijective encoding; and c) less obviously, the implementation call for a Chebyshev distance. Equation 6 and Figure 9 of our original report[3], reproduced here as Figure 5. The corresponding equation, reproduced below as Equation 2, calculates the similar length, *Sn* (in nucleotide units) between two positions as a logarithm function of the maximum distance along any axis (Chebyshev).

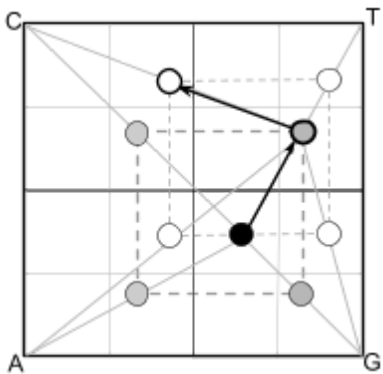

**Figure 5** - Graphic demonstration of how a location (and corresponding sequence) in the unit iterated map, can only be followed by the transition to 4 possible nucleotides at a Chebyshev distance of ½ from each other (the vertices of the dashed line square). The metric distance described in Equation 2 will accordingly produce a distance of $Sn = -log_2(½) = 1$ for the emission of any two distinct nucleotides. Accordingly, an aligned sequence will be reflected by consecutive halving. The figure illustrates this conservation for a sequence "GTC" starting at position ● and ending at ○ (arrows).

$$Sn^{direction}(u_i^{direction}, u_j^{direction}) \geq -log_2\left(\max\left|u_i^{direction} - u_j^{direction}\right|\right)$$

Equation 2

The underlying rationale becomes clear by noting that the next position in a sequence will occupy one of four possible 4 corners of a square with size ½ x ½. As recalled in the Introduction, this metric was updated to consider both forward and backward embeddings of the USM procedure[6], such that the addition does not underestimate the length of the similar segment (Equation 3). Note the -1 to avoid double counting

$$Sn^{forward} + Sn^{backward} - 1 \geq Sn*$$ 

Equation 3,
(using *Sn\** as defined in Eq.2)

It took a decade before an efficient packaging of USM coordinates became available as map-reduce operations[7]. It is to that implementation that the binning in Figure 4 owes the ability to process the ¼ million nucleotide sequence in the 10 ms range using a regular laptop.That report also found that an exact sequence similarity metric (the length of the similar segment, *Sn*) could be derived, still without alignment (Equation 6). Specifically, no other data about two units in a sequence is required to determine similar length beyond the forward and backward numerical embedding coordinates. An interactive tool accompanying that report, at usm.github.io, demonstrates the calculation by cross-tabulating Sn values for sequences of any alphabet. By devising a similarity metric, *Sn*, based on the number of quadrant foldings that two positions share, the overfitting is removed, and an exact, fully parallel implementation was found (Equation 6):

For each direction, *forward* and *backward*, and for each dimension, $Sn^{direction}$ is calculated independently for each unit and each dimension, *j*:

$$Sn^{direction}(u1, u2) = \max \begin{vmatrix} x = 0 \\ while\left(round(u1^{direction}_{j=1\ldots h} \cdot 2^x) == round(u2^{direction}_{j=1\ldots h} \cdot 2^x)\right)\{x = x+1\} \\ return\ x \end{vmatrix}$$

<div align="center">Equation 5</div>

causing the overall similar length, *Sn* between two embedded units, *u1* and *u2* to be exactly

$$Sn = Sn^{forward} + Sn^{backward} - 1 \qquad \text{Equation 6}$$
<div align="right">(using *Sn* as defined in Eq.5)</div>

Figure 6 revisits that interactive tool at usm.github.io (click on button "demo fold") to recall the effect of applying the similar length metric, *Sn*, described in Equation 6, to the cross-tabulation of two sequences, which for convenience we'll describe as a reference sequence and a prompt sequence. Importantly, each computation of *Sn*, is completely parallelized: filling each cell in the distance table needs only two inputs: the forward and the backward embedding coordinates for one unit in one sequence, and the forward and backward coordinates for another unit, in another sequence. That is, the computation of *Sn* is positionally agnostic. Scrambling or removing rows and columns will not per se change the value of the similar length, Sn, for any two remaining units, assessed from its embedded coordinates.

Figure 6 illustrates this process for the analysis of a base sequence "AAT**GATTACA**GGG" compared with a probing sequence (prompt) "TGA**GATTACA**CGTCA". The interactive tool also allows collecting individual coordinates by hovering the values, in red and blue, and using them to reassemble the raw sequence from the position where they were encoded. Finally, the github-hosted live application accompanying this report includes an interactive panel where the coordinates encoded/decoded.d $7 + 4 - 2 = 9$ (similarity of first subsequence + similarity of second subsequence - length of overlap). Since this calculation is focused on longest exact matches, then the overall similar distance would be 7. If a metric is needed to determine overall similarity between two sequences one might, for example, choose to weigh all scores equally, which in this case is 120.

| PROBE | forward | Map | [0.59, 0.70] | [0.79, 0.35] | [0.39, 0.17] | [0.69, 0.08] | [0.34, 0.04] | [0.67, 0.52] | [0.83, 0.76] | [0.41, 0.38] | [0.20, 0.69] | [0.10, 0.34] | [0.05, 0.67] | [0.52, 0.33] | [0.76, 0.66] | [0.38, 0.83] | [0.19, 0.41] | |
|---|---|---|---|---|---|---|---|---|---|---|---|---|---|---|---|---|---|---|
| BASE | backward | Map | [0.83, 0.52] | [0.67, 0.05] | [0.34, 0.10] | [0.69, 0.20] | [0.38, 0.41] | [0.76, 0.83] | [0.52, 0.66] | [0.05, 0.33] | [0.10, 0.67] | [0.20, 0.35] | [0.40, 0.70] | [0.80, 0.40] | [0.60, 0.81] | [0.20, 0.63] | [0.41, 0.26] | Reduce to: |
| forward | backward | seq | T | G | A | G | A | T | T | A | C | A | C | G | T | C | A | Sum | Max |
| [0.44,0.02] | [0.21,0.15] | A | 0 | 0 | 2 | 0 | 2 | 0 | 0 | 1 | 0 | 1 | 0 | 0 | 0 | 0 | 1 | 7 | 2 |
| [0.22,0.01] | [0.42,0.30] | A | 0 | 0 | 1 | 0 | 2 | 0 | 0 | 1 | 0 | 1 | 0 | 0 | 0 | 0 | 4 | 9 | 4 |
| [0.61,0.50] | [0.84,0.60] | T | 4 | 0 | 0 | 0 | 0 | 2 | 1 | 0 | 0 | 0 | 0 | 0 | 1 | 0 | 0 | 8 | 4 |
| [0.80,0.25] | [0.69,0.20] | G | 0 | 4 | 0 | 7 | 0 | 0 | 0 | 0 | 0 | 0 | 0 | 0 | 1 | 0 | 0 | 12 | 7 |
| [0.40,0.12] | [0.38,0.40] | A | 0 | 0 | 4 | 0 | 7 | 0 | 0 | 1 | 0 | 1 | 0 | 0 | 0 | 0 | 2 | 15 | 7 |
| [0.70,0.56] | [0.77,0.81] | T | 2 | 0 | 0 | 0 | 0 | 7 | 1 | 0 | 0 | 0 | 0 | 0 | 1 | 0 | 0 | 11 | 7 |
| [0.85,0.78] | [0.55,0.62] | T | 1 | 0 | 0 | 0 | 0 | 1 | 7 | 0 | 0 | 0 | 0 | 0 | 1 | 0 | 0 | 10 | 7 |
| [0.42,0.39] | [0.11,0.25] | A | 0 | 0 | 1 | 0 | 1 | 0 | 0 | 7 | 0 | 2 | 0 | 0 | 0 | 0 | 1 | 12 | 7 |
| [0.21,0.69] | [0.22,0.50] | C | 0 | 0 | 0 | 0 | 0 | 0 | 0 | 0 | 7 | 0 | 2 | 0 | 0 | 2 | 0 | 11 | 7 |
| [0.10,0.34] | [0.45,0.00] | A | 0 | 0 | 2 | 0 | 1 | 0 | 0 | 1 | 0 | 7 | 0 | 0 | 0 | 0 | 2 | 13 | 7 |
| [0.55,0.17] | [0.90,0.01] | G | 0 | 1 | 0 | 2 | 0 | 0 | 0 | 0 | 0 | 0 | 0 | 1 | 0 | 0 | 0 | 4 | 2 |
| [0.77,0.08] | [0.80,0.03] | G | 0 | 1 | 0 | 1 | 0 | 0 | 0 | 0 | 0 | 0 | 0 | 1 | 0 | 0 | 0 | 3 | 1 |
| [0.88,0.04] | [0.60,0.07] | G | 0 | 2 | 0 | 2 | 0 | 0 | 0 | 0 | 0 | 0 | 0 | 1 | 0 | 0 | 0 | 5 | 2 |
| | | | | | | | | | | | | | | | | | | 120 | 7 |

**Figure 6** - Cross-tabulation of *Sn* similarity metric (Equation 6) between all elements of "AAT**GATTACA**GGG" and "TGA**GATTACA**CGTCA" using circular seeding (Figure 2). This table was produced using the web tool at usm.github.io [7], and can be reproduced by clicking on the green button "demo Fold". Note how the circular encoding causes the last column of the score table to continue as its first column. In this example, the longest similar subsegment, length 7, was found to be the sequence pattern "GATTACA", and the second longest is "ATGA", with a similar length of 4. The last base of sequence 2 is an "A", which due to circular seeding links up with the "TGA" at the start of the sequence.

An overlap between the two similar segments is apparent in the figure 6, yellow highlight, where the first two nucleotides of the longer similar segment, GA, are the same last two nucleotides of the shorter similar segment, together maybe suggesting a composite similarity metric defined as the length $7 + 4 - 2 = 9$ (similarity of first subsequence + similarity of second subsequence - length of overlap). Since the calculation in Figure 6 is focused on the longest exact match, the overall similar distance is 7. If a different metric is needed, for example, to determine overall similarity between two sequences one might choose to weigh all scores equally, which in this case is 120.

- C1) Pick a position from the reference sequence 12137  (pick from reference)
...CTTAAAGCTCCCTTGGCAATTCTGAGGAGT[G]→G←[A]TTACATGTTGTATGTAGCTCGTAACGAAA...
- C2) Embedding coordinates (maybe start by picking from training data, above, and then try with your own sequences, below)
Forward - 1 = [G]( 0.92849555931755765 , 0.2537002004294902 ): bin [ 237 , 64 ] = 5
Backward +1 = [A]( 0.39049280485731175 , 0.417316094009723014 ): bin [ 99 , 106 ] = 0

**Figure 7** - Snapshot of web tool displaying bidirectional embedding converging on the identity of the forward encoded head token, and the backward encoded tail token. If this web tool (see Availability) uses the reference EGFR sequence and is pointed to position 12137, the convergence points to "G". Clicking on "pick from reference will activate interactive bidirectional mapping.

```
ia = ▼usm {
  seq: "MRPSGTAGAALLALLAALCPASRALEEKKVCQGTSNKLTQLGTFEDHFLSLQRMFNNCEVVLGNLEITYVQRNYDLSFLKTIQEVAGYVLIALNTVERIP"
  edges: ▼Object {
    A: ▶Array(5) [0, 0, 0, 0, 0]
    C: ▶Array(5) [0, 0, 0, 0, 1]
    D: ▶Array(5) [0, 0, 0, 1, 0]
    E: ▶Array(5) [0, 0, 0, 1, 1]
    F: ▶Array(5) [0, 0, 1, 0, 0]
    G: ▶Array(5) [0, 0, 1, 0, 1]
    H: ▶Array(5) [0, 0, 1, 1, 0]
    I: ▶Array(5) [0, 0, 1, 1, 1]
    K: ▶Array(5) [0, 1, 0, 0, 0]
    L: ▶Array(5) [0, 1, 0, 0, 1]
    M: ▶Array(5) [0, 1, 0, 1, 0]
    N: ▶Array(5) [0, 1, 0, 1, 1]
    P: ▶Array(5) [0, 1, 1, 0, 0]
    Q: ▶Array(5) [0, 1, 1, 0, 1]
    R: ▶Array(5) [0, 1, 1, 1, 0]
    S: ▶Array(5) [0, 1, 1, 1, 1]
    T: ▶Array(5) [1, 0, 0, 0, 0]
    V: ▶Array(5) [1, 0, 0, 0, 1]
    W: ▶Array(5) [1, 0, 0, 1, 0]
    Y: ▶Array(5) [1, 0, 0, 1, 1]
  }
  forward: ▼Array(5) [
    0: ▶Array(100) [0.007812500495731456, 0.003906250247865728, 0.001953125123932864, 0.000976562561966432, 0.0…
    1: ▶Array(100) [0.9691645681439258, 0.9845822840719629, 0.9922911420359815, 0.9961455710179907, 0.498072785…
    2: ▶Array(100) [0.23632878084087022, 0.618164390420435, 0.8090821952102175, 0.9045410976051087, 0.952270548…
    3: ▶Array(100) [0.9062501900116585, 0.9531250950058292, 0.4765625475029146, 0.7382812737514572, 0.369140636…
    4: ▶Array(100) [0.049243119511531830, 0.024621559755765916, 0.012310779877882958, 0.5061553899389415, 0.7530…
  ]
  backward: ▼Array(5) [
    0: ▶Array(100) [0.015625000991462912, 0.031250001982925824, 0.06250000396585165, 0.1250000079317033, 0.2500…
    1: ▶Array(100) [0.9383291362878516, 0.8766582725757033, 0.7533165451514064, 0.5066330903028129, 0.013266180…
    2: ▶Array(100) [0.47265756168174045, 0.9453151233634809, 0.8906302467269619, 0.7812604934539238, 0.56252098…
    3: ▶Array(100) [0.812500380023317, 0.6250007600466341, 0.25000152009326804, 0.5000030401865361, 0.000006080…
    4: ▶Array(100) [0.09848623902306367, 0.19697247804612733, 0.39394495609225466, 0.7878899121845093, 0.575779…
  ]
  canvas: f(…)
}
```

seq="MRPSGTAGAALLALLAALCPASRALEEKKVCQGTSNKLTQLGTFE
DHFLSLQRMFNNCEVVLGNLEITYVQRNYDLSFLKTIQEVAGYVLIALNT
VERIP"

aa="ACDEFGHIKLMNPQRSTVWY"

ua = new usm(seq, aa)

Alphabet, aa
(20 aminoacids, note this causes h=5)

bidirectional embedded numerical space

https://observablehq.com/@episphere/usm

**Figure 8** - embedding of the first 100 aminoacids of the EGFR protein sequence. Note how adding additional dimensions to the unit hypercube in order to accommodate longer alphabets does not require changing the encoding procedure. The same is the case for the projection in the frequency domain, FCGR.

# Discussion

### USM as a language model

In its simplest form, a language model is a function, *P*, that returns a probability of the emission of a $T_n$ token when provided with the sequence of *n-1* past tokens, $P(T_n | [T_1, ... , T_{n-1}])$. The bidirectional nature of the USM encoder (figure 8) advances this formulation by calculating a simultaneous forward and a backward density distribution (Figure 4). These frequency tables have been used as input for image analysis classification [8] such as convolutional neural networks and variations of FCGR[25] that compare favorably with computational intensive DNABERT[18,25].

The accompanying Fold web application includes an interactive tool to visualize how the three components of USM articulate with each other (Figure 7): by moving the range button one can inspect how sequence identity is encoded in an embedded numeric space bidirectionally and bijectively, and also how binning USM coordinates by quadrant computes the corresponding frequency domain representation efficiently. As a consequence, the third component of this architecture is now in place, providing a framework to explore the association of regions of the embedded space with the probability of emitting candidate tokens. In its simplest form such a model could be an n-gram with the transition probabilities provided by USM's frequency space (FCGR, Figure 4). A different approach may instead use the *Sn* distance matrix (Figure 6) as the starting point for a generative modelling.

USM embeddings find straightforward applications to Biology, such as the use of mutation signatures (k-mer frequencies) to track somatic evolution in cancer etiology. Another avenue worth exploring might have advancements in Bioinformatics algorithms using bijective encoding with operation in fractal state-spaces. However, the focus of this report is circumscribed to the numeric methods behind the extraordinary properties of iterated Universal Sequence maps. Among those, the bijective similarity metric that can be calculated order-free stands particularly high. Finally, the one wrinkle in the USM map assembly caused by ignoring its nature as the convergence of a numeric process was solved by revisiting its roots - the seeding of that iterative process.

# Conclusion

Iterated Universal Sequence Maps (USM) provide a 1) bijective encoder; 2) computationally efficient through parallelization; 3) scale-free, k-mer frequencies can be recalculated for different k's from embedding vectors; 4) order-free, the calculation of similar length distance metric, Sn, only requires the corresponding two sets of embedded coordinates; 5) a generic mapping between symbolic sequences and numeric spaces, i.e. latent embedding space. The iterated procedure can be applied to any sequence with a finite alphabet, from genomic sequences to protein conformations and natural language words. In this study, we analyse the components of USM needed to generate numeric vectors that non-linear classifiers can use as training inputs for both pattern discovery and recognition.The main finding was that USM is best approached as a numeric process where the feature vectors are determined by convergence triggered by a suitable numeric seeding mechanism. It is important to clarify that, although USM could be treated as an encode/decode language model, the research direction proposed here is focused on the statistical regularities of the fractal embedded space.

# References

1. Jeffrey HJ. Chaos game representation of gene structure. Nucleic Acids Res. 1990;18: